\documentclass{article}

\usepackage{PRIMEarxiv}

\usepackage[utf8]{inputenc} 
\usepackage[T1]{fontenc}    
\usepackage{hyperref}       
\usepackage{url}            
\usepackage{booktabs}       
\usepackage{amsfonts}       
\usepackage{nicefrac}       
\usepackage{microtype}      
\usepackage{lipsum}
\usepackage{fancyhdr}       
\usepackage{graphicx}       
\usepackage{amsmath}
\graphicspath{{media/}}     

\usepackage{pgfplots}
\pgfplotsset{compat=1.7}

\title{Deep Linear Discriminant Analysis with Variation for Polycystic Ovary Syndrome Classification}

\author{
    Raunak Joshi \\
      Department of IT \\
      University of Mumbai \\
      Mumbai, 400032, India \\
      \texttt{raunakjoshi.m@gmail.com} \\
    \And
  Abhishek Gupta \\
  Department of EXTC \\
  University of Mumbai \\
  Mumbai, 400032, India \\
  \texttt{abhishekgupta20001@gmail.com} \\
   \And
  Himanshu Soni \\
  Department of EXTC \\
  SJCEM \\
  Palghar, 401404, India \\
  \texttt{himansh056@gmail.com} \\
  \And
  \AND
  Ronald Laban \\
  Department of EXTC \\
  SJCEM \\
  Palghar, 401404, India \\
  \texttt{ronaldlaban@gmail.com} \\
}

\begin{document}
\maketitle

\begin{abstract}
The polycystic ovary syndrome diagnosis is a problem that can be leveraged using prognostication based learning procedures. Many implementations of PCOS can be seen with Machine Learning but the algorithms have certain limitations in utilizing the processing power graphical processing units. The simple machine learning algorithms can be improved with advanced frameworks using Deep Learning. The Linear Discriminant Analysis is a linear dimensionality reduction algorithm for classification that can be boosted in terms of performance using deep learning with Deep LDA, a transformed version of the traditional LDA. In this result oriented paper we present the Deep LDA implementation with a variation for prognostication of PCOS.
\end{abstract}


\section{Introduction}
The use of medical research data for various statistical tasks has been done from a prolonged period of time. The data after having consistent number of records can be utilized for deriving inference using prognostication methods. The methods that fall under the area of inferential statistics \cite{Marshall2011AnIT} which are extended with applied areas of statistics can be used, one of which can be used is Machine Learning \cite{sarker2021machine}. Task of prognostication based on data can be done by learning patterns from the data using machine learning. The dataset that we have used in this paper is pertaining to polycystic ovary syndrome \cite{ndefo2013polycystic} diagnosis, which falls under the classification \cite{10.2307/2344237} category. Classification has 2 sub-divisions, viz. Binary and Multi-Class where the data used in this paper falls under binary classification \cite{10.5120/ijca2017913083} precisely. The methods using Machine Learning have already been performed on polycystic ovary syndrome abbreviated as PCOS using logistic regression \cite{9510128}, bagging ensemble methods  \cite{kanvinde2022binary}, discriminant analysis \cite{gupta2022discriminant}, stacked generalization \cite{nair2022combining}, boosting ensemble methods \cite{gupta2021succinct} and deep neural networks \cite{gupta2022residual}. The use of deep learning is evident and we want to focus on the more variations that can be brought into the current state-of-the-art system. Deep Learning \cite{lecun2015deep,goodfellow2016deep} provides more depth of learning as compared to machine learning and is used when the amount of parameters in dimensions are high. The PCOS dataset has over 41 dimensions which are enough for instating the use of deep learning. The variation we wanted to perform was related to some machine learning algorithm that can be leveraged with power of deep learning. The implementation of any machine learning algorithm using a library like scikit-learn \cite{scikit-learn,sklearn_api} meets limitations in terms of utilization with GPU processing power. For the same reason, using a mature framework like Tensorflow \cite{tensorflow2015-whitepaper} can definitely bring change to the working. This is where we decided to work with parametric learning method which works with simple and definite procedures. The parametric learning method we focused on using was discriminant analysis \cite{10.2307/143150,10.1007/978-1-4612-6079-0_16} which has variations in it where we focused on Linear Discriminant Analysis \cite{Tharwat2017LinearDA}. This actually accounted for an idea that training the linear discriminant analysis with deep learning style will yield us Deep Linear Discriminant Analysis \cite{dorfer2015deep,8014812} which has been already been discovered and we decided to proceed with out implementation using it. The variations that we brought in the network will be explained in further sections of this paper. 

\section{Methodology}
This section of the paper gives detailed insights about the implementation and approach we have taken to solve the problem. The model considering the implementation with respect to Deep LDA \cite{dorfer2015deep,8014812} revolves around the idea of the convolutional neural networks \cite{lecun1995convolutional,alzubaidi2021review,gupta2022detection,joshi2022refactoring}. The modification can be done to work with numerical values. This has been implemented by various developers and names of the developers are given in the acknowledgement section of the paper. The Deep LDA is basically an implementation of latent representations in linearly separable method. The Deep LDA is an extensive implementation of the traditional Linear Discriminant Analysis which was intended for dimensionality reduction based classification methods. The implementation consists of 2 phases, first phase consists of linear discriminator as the deep neural network and second phase consists of support vector machine for detailed classification.

\subsection{First Phase}

\begin{figure}[htbp]
    \centering
    \includegraphics[angle=270,scale=0.9]{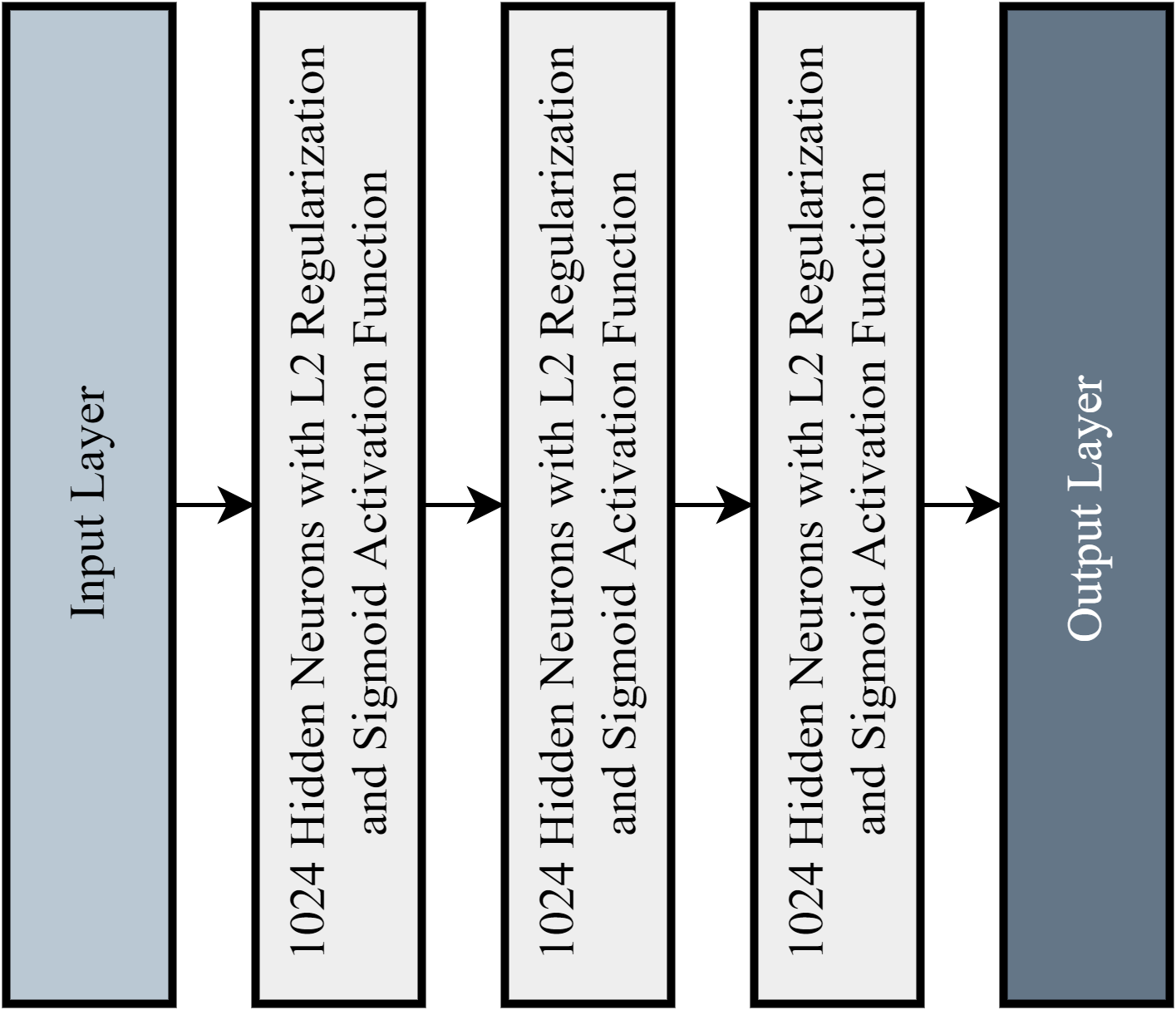}
    \caption{First Phase with LDA Implementation}
    \label{fig:a}
\end{figure}

The figure \ref{fig:a} gives depiction of first phase of the implementation. The input layer takes 41 dimension of features from the data. This is passed on to one dense layer that has 1024 hidden neurons. The L2 regularization \cite{cortes2012l2} is applied as a kernel regularization for the layer. The activation function is used sigmoid \cite{10.1016/S0893-6080(05)80129-7}. The rectified linear unit abbreviated as ReLU \cite{agarap2018deep} is the activation function commonly used for deep learning methods but using sigmoid ensures linear based system such as developed for linear discriminant analysis. The parameters learned from the first hidden layer are 43,008. Similar type of hidden layers are repeated twice, where second hidden layer learns 1,049,600 parameters and third hidden layer also learns same amount of parameters. The output layer consists of 1 hidden neuron with sigmoid activation function and learns 1025 parameters. The network learns total of 2,143,233 parameters where all the parameters are trainable. The loss function used is binary cross-entropy that differs from the original implementation of deep LDA paper. The loss optimizer used is Adam \cite{kingma2014adam} optimizer with learning rate of $1*10^{-5}$ that roughly denotes 0.00001. The implementation is done using Keras \cite{chollet2015keras} over Tensorflow \cite{tensorflow2015-whitepaper} back-end trained for 100 epochs with 64 as batch size.

\subsection{Second Phase}

\begin{figure}[htbp]
    \centering
    \includegraphics[angle=270,scale=0.9]{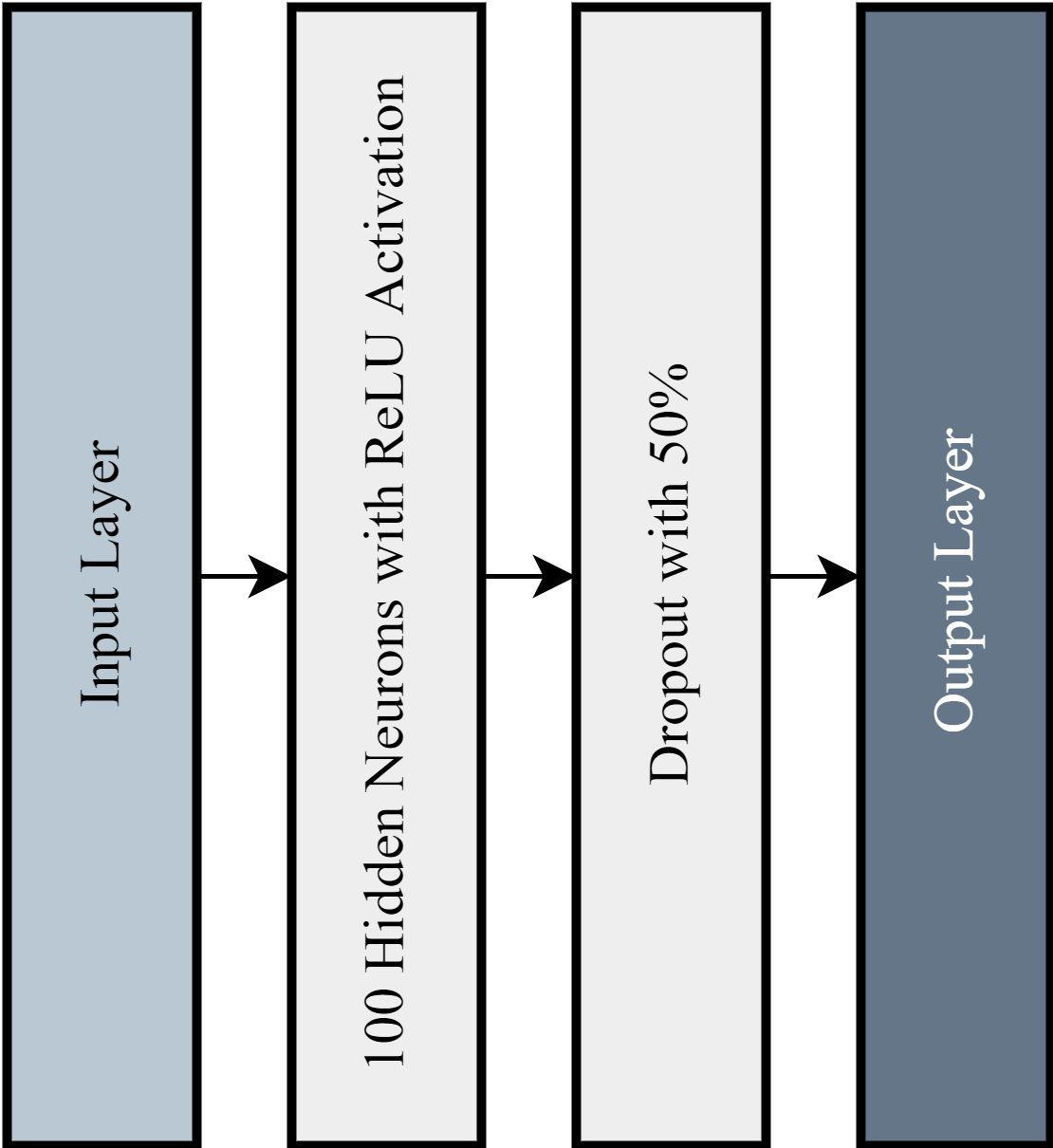}
    \caption{Second Phase with SVM Implementation}
    \label{fig:b}
\end{figure}

The figure \ref{fig:b} depicts the second phase implementation. This is done using Support Vector Machine \cite{708428} implementation with neural network inclination. The input layer is connected to hidden layer with 100 hidden neurons with ReLU \cite{agarap2018deep} activation function. This layer learns 200 parameters. This layer is connected to dropout \cite{JMLR:v15:srivastava14a} layer with 50\% threshold and does not learn any parameters. The output layer has 1 hidden neuron and has sigmoid activation function for binary classification and learns 101 parameters. The total parameters learned are 301 and the network uses binary cross-entropy. The loss optimizer used is Adam \cite{kingma2014adam} optimizer with  $1*10^{-5}$ that approximately denotes 0.00001 learning rate. The network is trained with 100 epochs and 64 as batch size.

\subsection{Complete Network}

\begin{figure}[htbp]
    \centering
    \includegraphics[angle=270,scale=0.8]{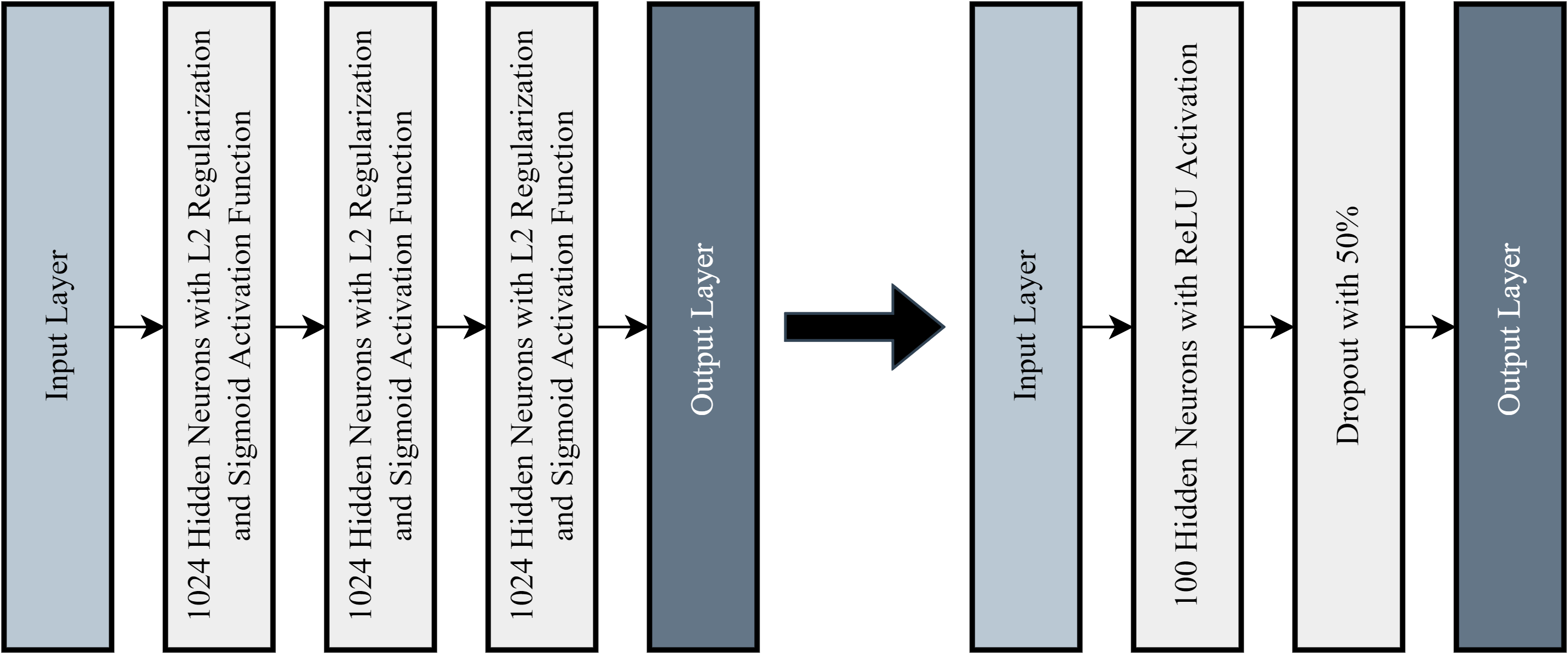}
    \caption{Complete Network}
    \label{fig:c}
\end{figure}

The complete network is accumulation of first and second phase where the output of the first phase is the input for second phase. The output of the first phase is 1-dimensional array from 41-dimensional output. This is given as an input to the second phase of the network and final prediction which is 1-dimensional is achieved. The both phases are trained independently and the output is retained from first phase and given as second phase. The results of the network will be given in succeeding section of the paper.

\section{Results}
\subsection{Accuracy and Loss for First Phase}

\begin{figure}[htbp]
    \centering
    \begin{tabular}{c c}
        \begin{tikzpicture}[scale=0.6]
            \begin{axis}[xlabel={Epochs},ylabel ={Accuracies},enlargelimits=false,
            grid=both,
            scale only axis=true,legend pos=south east,style={ultra thick}, axis line style={ultra thin}]
            \addplot+[no markers] table[x=Epochs,y=accuracy,col sep=comma]{plots/lda.csv}; 
            \addplot+[no markers] table[x=Epochs,y=val_accuracy,col sep=comma]{plots/lda.csv};
            \addlegendentry{Training Accuracy}
            \addlegendentry{Validation Accuracy}
        \end{axis}
        \end{tikzpicture}
        &
        \begin{tikzpicture}[scale=0.6]
            \begin{axis}[xlabel={Epochs},ylabel ={Losses},enlargelimits=false,
            grid=both,
            scale only axis=true,legend pos=north east,style={ultra thick}, axis line style={ultra thin}]
            \addplot+[no markers] table[x=Epochs,y=loss,col sep=comma]{plots/lda.csv}; 
            \addplot+[no markers] table[x=Epochs,y=val_loss,col sep=comma]{plots/lda.csv};
            \addlegendentry{Training Loss}
            \addlegendentry{Validation Loss}
        \end{axis}
        \end{tikzpicture}
    \end{tabular}
    \caption{Accuracy and Loss for the First Phase}
    \label{fig:d}
\end{figure}
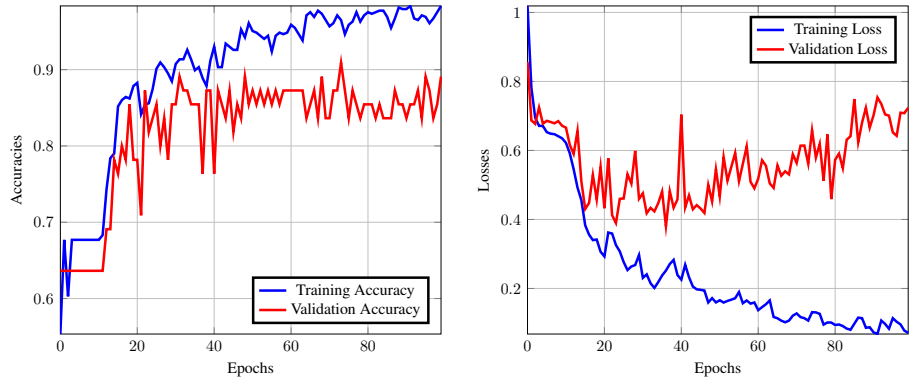

The training and validation accuracy graph can be seen from figure \ref{fig:d} and the infractions between the training and validation accuracy seem a bit wider but the values for training are 98.35\% and validation 90.909\% respectively. Considering the loss, the validation loss has some infraction that can be seen with different metrics the training loss is 6.79\% whereas the validation loss is 38.05\% respectively.

\subsection{Accuracy and Loss for Second Phase}
The second phase contains support vector machine and not necessarily the graph depiction is right measure but we have included the graph. The inference can be drawn as there are no significant changes in the accuracy or loss. All the metrics are learnt from the trained parameters of the first phase and it does not make much sense to make mappings out of it. The training accuracy for the support vector machine phase is generated as 98.354\% and validation accuracy is obtained as 90.909\% which is similar to the first phase training and validation accuracy. The training loss is 6.79\% and validation loss is 38.052\% which is again similar to the first phase. The better inference can be generated from different metrics intended for classification and not just the graph depictions of the training and validation accuracy as well as loss.

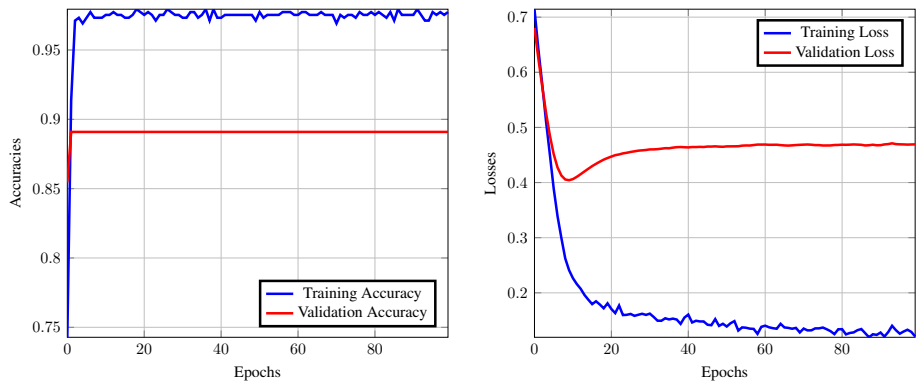
\begin{figure}[htbp]
    \centering
    \begin{tabular}{c c}
        \begin{tikzpicture}[scale=0.6]
            \begin{axis}[xlabel={Epochs},ylabel ={Accuracies},enlargelimits=false,
            grid=both,
            scale only axis=true,legend pos=south east,style={ultra thick}, axis line style={ultra thin}]
            \addplot+[no markers] table[x=Epochs,y=accuracy,col sep=comma]{plots/svm.csv}; 
            \addplot+[no markers] table[x=Epochs,y=val_accuracy,col sep=comma]{plots/svm.csv};
            \addlegendentry{Training Accuracy}
            \addlegendentry{Validation Accuracy}
        \end{axis}
        \end{tikzpicture}
        &
        \begin{tikzpicture}[scale=0.6]
            \begin{axis}[xlabel={Epochs},ylabel ={Losses},enlargelimits=false,
            grid=both,
            scale only axis=true,legend pos=north east,style={ultra thick}, axis line style={ultra thin}]
            \addplot+[no markers] table[x=Epochs,y=loss,col sep=comma]{plots/svm.csv}; 
            \addplot+[no markers] table[x=Epochs,y=val_loss,col sep=comma]{plots/svm.csv};
            \addlegendentry{Training Loss}
            \addlegendentry{Validation Loss}
        \end{axis}
        \end{tikzpicture}
    \end{tabular}
    \caption{Accuracy and Loss for the Second Phase}
    \label{fig:e}
\end{figure}

\subsection{Precision}
The precision \cite{powers2020evaluation} is dependent on total number of samples that are predicted to be positive among all the set of samples. This is a popular metric for prognostication algorithms and requires basic elements of a confusion matrix \cite{ting2017confusion}, viz. true positives, true negatives, false positives and false negatives. The precision score obtained in 88.88\% which is very close to 1 as expected.

\subsection{Recall}
The recall \cite{powers2020evaluation} is a metrics which states all the positive elements from the every single predicted element. The recall also utilizes every single element from the confusion matrix same as precision. The recall generated for the model is 80\% which is again a very good score and gives inference that model has performed adequately. The recall is not only metric that gives the final inference and more precise metric can be obtained.

\subsection{F-Score}
This is a subtle metric that gives the overall flow of how efficiently does the model perform. The building blocks of F-score \cite{powers2020evaluation,goutte2005probabilistic} are precision and recall. The F-Score we got for the model is 84.21\%, which is adequately good and proves that model performs better on average.

\section{Conclusion}
The idea of entire paper revolves around an experimentation that can be performed for polycystic ovary syndrome diagnosis problem. We proved a point that simple machine learning algorithms can be leveraged using deep learning for efficient performance based inclination. The Deep Linear Discriminant Analysis idea was proven in this paper. We also introduced some of our personal variations in implementation and they turned out to be effective from the results section of the paper. The paper can definitely sum up many lost ideas into practical implementation and enforce many young researchers for better development perspective.

\section*{Acknowledgements}
We would genuinely like to thank Mr. Prasoon Kottarathil for making the polycystic ovary syndrome dataset available through Kaggle platform. We would also like to deeply thank for the contributions of VahidooX - \emph{\textbf{https://github.com/VahidooX/DeepLDA}} and Thomas Chaton - \emph{\textbf{https://github.com/tchaton/DeepLDA}} for providing us the basic implementations of Deep LDA based from the original paper.

\bibliographystyle{unsrt}  
\bibliography{references}

\begin{thebibliography}{10}

\bibitem{Marshall2011AnIT}
Gill Marshall and Leon Jonker.
\newblock An introduction to inferential statistics: A review and practical
  guide.
\newblock {\em Radiography}, 17, 2011.

\bibitem{sarker2021machine}
Iqbal~H Sarker.
\newblock Machine learning: Algorithms, real-world applications and research
  directions.
\newblock {\em SN Computer Science}, 2(3):1--21, 2021.

\bibitem{ndefo2013polycystic}
Uche~Anadu Ndefo, Angie Eaton, and Monica~Robinson Green.
\newblock Polycystic ovary syndrome: a review of treatment options with a focus
  on pharmacological approaches.
\newblock {\em Pharmacy and therapeutics}, 38(6):336, 2013.

\bibitem{10.2307/2344237}
R.~M. Cormack.
\newblock A review of classification.
\newblock {\em Journal of the Royal Statistical Society. Series A (General)},
  134(3):321--367, 1971.

\bibitem{10.5120/ijca2017913083}
Roshan Kumari and Saurabh~Kr. Srivastava.
\newblock Machine learning: A review on binary classification.
\newblock {\em International Journal of Computer Applications}, 160(7):11--15,
  Feb 2017.

\bibitem{9510128}
Preeti Chauhan, Pooja Patil, Neha Rane, Pooja Raundale, and Harshil Kanakia.
\newblock Comparative analysis of machine learning algorithms for prediction of
  pcos.
\newblock In {\em 2021 International Conference on Communication information
  and Computing Technology (ICCICT)}, pages 1--7, 2021.

\bibitem{kanvinde2022binary}
Nandan Kanvinde, Abhishek Gupta, and Raunak Joshi.
\newblock Binary classification for high dimensional data using supervised
  non-parametric ensemble method.
\newblock {\em arXiv preprint arXiv:2202.07779}, 2022.

\bibitem{gupta2022discriminant}
Abhishek Gupta, Himanshu Soni, Raunak Joshi, and Ronald~Melwin Laban.
\newblock Discriminant analysis in contrasting dimensions for polycystic ovary
  syndrome prognostication.
\newblock {\em arXiv preprint arXiv:2201.03029}, 2022.

\bibitem{nair2022combining}
Sruthi Nair, Abhishek Gupta, Raunak Joshi, and Vidya Chitre.
\newblock Combining varied learners for binary classification using stacked
  generalization.
\newblock {\em arXiv preprint arXiv:2202.08910}, 2022.

\bibitem{gupta2021succinct}
Abhishek~M Gupta, Sannidhi~S Shetty, Raunak~M Joshi, and Ronald~Melwin Laban.
\newblock Succinct differentiation of disparate boosting ensemble learning
  methods for prognostication of polycystic ovary syndrome diagnosis.
\newblock In {\em 2021 International Conference on Advances in Computing,
  Communication, and Control (ICAC3)}, pages 1--5. IEEE, 2021.

\bibitem{gupta2022residual}
Abhishek Gupta, Sruthi Nair, Raunak Joshi, and Vidya Chitre.
\newblock Residual-concatenate neural network with deep regularization layers
  for binary classification.
\newblock {\em arXiv preprint arXiv:2205.12775}, 2022.

\bibitem{lecun2015deep}
Yann LeCun, Yoshua Bengio, and Geoffrey Hinton.
\newblock Deep learning.
\newblock {\em nature}, 521(7553):436--444, 2015.

\bibitem{goodfellow2016deep}
Ian Goodfellow, Yoshua Bengio, and Aaron Courville.
\newblock {\em Deep learning}.
\newblock 2016.

\bibitem{scikit-learn}
F.~Pedregosa, G.~Varoquaux, A.~Gramfort, V.~Michel, B.~Thirion, O.~Grisel,
  M.~Blondel, P.~Prettenhofer, R.~Weiss, V.~Dubourg, J.~Vanderplas, A.~Passos,
  D.~Cournapeau, M.~Brucher, M.~Perrot, and E.~Duchesnay.
\newblock Scikit-learn: Machine learning in {P}ython.
\newblock {\em Journal of Machine Learning Research}, 12:2825--2830, 2011.

\bibitem{sklearn_api}
Lars Buitinck, Gilles Louppe, Mathieu Blondel, Fabian Pedregosa, Andreas
  Mueller, Olivier Grisel, Vlad Niculae, Peter Prettenhofer, Alexandre
  Gramfort, Jaques Grobler, Robert Layton, Jake VanderPlas, Arnaud Joly, Brian
  Holt, and Ga{\"{e}}l Varoquaux.
\newblock {API} design for machine learning software: experiences from the
  scikit-learn project.
\newblock In {\em ECML PKDD Workshop: Languages for Data Mining and Machine
  Learning}, pages 108--122, 2013.

\bibitem{tensorflow2015-whitepaper}
Mart\'{i}n Abadi, Ashish Agarwal, Paul Barham, Eugene Brevdo, Zhifeng Chen,
  Craig Citro, Greg~S. Corrado, Andy Davis, Jeffrey Dean, Matthieu Devin,
  Sanjay Ghemawat, Ian Goodfellow, Andrew Harp, Geoffrey Irving, Michael Isard,
  Yangqing Jia, Rafal Jozefowicz, Lukasz Kaiser, Manjunath Kudlur, Josh
  Levenberg, Dandelion Man\'{e}, Rajat Monga, Sherry Moore, Derek Murray, Chris
  Olah, Mike Schuster, Jonathon Shlens, Benoit Steiner, Ilya Sutskever, Kunal
  Talwar, Paul Tucker, Vincent Vanhoucke, Vijay Vasudevan, Fernanda Vi\'{e}gas,
  Oriol Vinyals, Pete Warden, Martin Wattenberg, Martin Wicke, Yuan Yu, and
  Xiaoqiang Zheng.
\newblock {TensorFlow}: Large-scale machine learning on heterogeneous systems,
  2015.
\newblock Software available from tensorflow.org.

\bibitem{10.2307/143150}
Leslie~J. King.
\newblock Discriminant analysis: A review of recent theoretical contributions
  and applications.
\newblock {\em Economic Geography}, 46:367--378, 1970.

\bibitem{10.1007/978-1-4612-6079-0_16}
Somesh Das~Gupta.
\newblock Discriminant analysis.
\newblock In Stephen~E. Fienberg and David~V. Hinkley, editors, {\em R.A.
  Fisher: An Appreciation}, pages 161--170, New York, NY, 1980. Springer New
  York.

\bibitem{Tharwat2017LinearDA}
Alaa Tharwat, Tarek Gaber, Abdelhameed Ibrahim, and Aboul~Ella Hassanien.
\newblock Linear discriminant analysis: A detailed tutorial.
\newblock {\em AI Commun.}, 30:169--190, 2017.

\bibitem{dorfer2015deep}
Matthias Dorfer, Rainer Kelz, and Gerhard Widmer.
\newblock Deep linear discriminant analysis.
\newblock {\em arXiv preprint arXiv:1511.04707}, 2015.

\bibitem{8014812}
Qing Tian, Tal Arbel, and James~J. Clark.
\newblock Deep lda-pruned nets for efficient facial gender classification.
\newblock In {\em 2017 IEEE Conference on Computer Vision and Pattern
  Recognition Workshops (CVPRW)}, pages 512--521, 2017.

\bibitem{lecun1995convolutional}
Yann LeCun, Yoshua Bengio, et~al.
\newblock Convolutional networks for images, speech, and time series.

\bibitem{alzubaidi2021review}
Laith Alzubaidi, Jinglan Zhang, Amjad~J Humaidi, Ayad Al-Dujaili, Ye~Duan,
  Omran Al-Shamma, Jos{\'e} Santamar{\'\i}a, Mohammed~A Fadhel, Muthana
  Al-Amidie, and Laith Farhan.
\newblock Review of deep learning: Concepts, cnn architectures, challenges,
  applications, future directions.
\newblock {\em Journal of big Data}, 8(1):1--74, 2021.

\bibitem{gupta2022detection}
Abhishek Gupta, Raunak Joshi, and Ronald Laban.
\newblock Detection of tool based edited images from error level analysis and
  convolutional neural network.
\newblock {\em arXiv preprint arXiv:2204.09075}, 2022.

\bibitem{joshi2022refactoring}
Raunak~M Joshi and Deven Shah.
\newblock Refactoring faces under bounding box using instance segmentation
  algorithms in deep learning for replacement of editing tools.
\newblock In {\em Intelligent Computing and Networking}, pages 236--247.
  Springer, 2022.

\bibitem{cortes2012l2}
Corinna Cortes, Mehryar Mohri, and Afshin Rostamizadeh.
\newblock L2 regularization for learning kernels.
\newblock {\em arXiv preprint arXiv:1205.2653}, 2012.

\bibitem{10.1016/S0893-6080(05)80129-7}
Ali~A. Minai and Ronald~D. Williams.
\newblock Original contribution: On the derivatives of the sigmoid.
\newblock {\em Neural Netw.}, 6(6):845–853, jun 1993.

\bibitem{agarap2018deep}
Abien~Fred Agarap.
\newblock Deep learning using rectified linear units (relu).
\newblock {\em arXiv preprint arXiv:1803.08375}, 2018.

\bibitem{kingma2014adam}
Diederik~P Kingma and Jimmy Ba.
\newblock Adam: A method for stochastic optimization.
\newblock {\em arXiv preprint arXiv:1412.6980}, 2014.

\bibitem{chollet2015keras}
Fran\c{c}ois Chollet et~al.
\newblock Keras.
\newblock \url{https://github.com/fchollet/keras}, 2015.

\bibitem{708428}
M.A. Hearst, S.T. Dumais, E.~Osuna, J.~Platt, and B.~Scholkopf.
\newblock Support vector machines.
\newblock {\em IEEE Intelligent Systems and their Applications}, 13(4):18--28,
  1998.

\bibitem{JMLR:v15:srivastava14a}
Nitish Srivastava, Geoffrey Hinton, Alex Krizhevsky, Ilya Sutskever, and Ruslan
  Salakhutdinov.
\newblock Dropout: A simple way to prevent neural networks from overfitting.
\newblock {\em Journal of Machine Learning Research}, 15(56):1929--1958, 2014.

\bibitem{powers2020evaluation}
David~MW Powers.
\newblock Evaluation: from precision, recall and f-measure to roc,
  informedness, markedness and correlation.
\newblock {\em arXiv preprint arXiv:2010.16061}, 2020.

\bibitem{ting2017confusion}
Kai~Ming Ting.
\newblock Confusion matrix.
\newblock {\em Encyclopedia of machine learning and data mining}, 260, 2017.

\bibitem{goutte2005probabilistic}
Cyril Goutte and Eric Gaussier.
\newblock A probabilistic interpretation of precision, recall and f-score, with
  implication for evaluation.
\newblock In {\em European conference on information retrieval}, pages
  345--359. Springer, 2005.

\end{thebibliography}

\end{document}